\documentclass[twocolumn,fontsize=9pt,paper=a4,pagesize,DIV=calc]{scrartcl}

\usepackage{graphicx}
\usepackage{tabularx}
\usepackage{array}
\usepackage{multirow}
\usepackage{amsmath}
\usepackage{amssymb}
\usepackage[version=3]{mhchem}
\usepackage[hidelinks, bookmarksopen]{hyperref}
\usepackage{bm}
\usepackage[T1]{fontenc}
\usepackage{xcolor}
\usepackage{xfrac}
\usepackage{subfig}
\usepackage{pythonhighlight}
\usepackage{framed}
\usepackage{dirtytalk}
\usepackage{tipa}

\definecolor{codegreen}{rgb}{0.5,0.5,0.8}
\definecolor{codegray}{rgb}{0.5,0.5,0.5}
\definecolor{codepurple}{rgb}{0,0.6,0}
\definecolor{backcolour}{rgb}{0.95,0.95,0.95}

\lstdefinestyle{mystyle}{
    backgroundcolor=\color{backcolour},   
    commentstyle=\color{codegreen},
    keywordstyle=\color{orange},
    numberstyle=\tiny\color{codegray},
    stringstyle=\color{codepurple},
    basicstyle=\ttfamily\footnotesize,
    breakatwhitespace=false,         
    breaklines=true,                 
    captionpos=b,                    
    keepspaces=true,                 
    numbers=left,                    
    numbersep=5pt,                  
    showspaces=false,                
    showstringspaces=false,
    showtabs=false,                  
    tabsize=2
}

\usepackage[backend=biber,style=phys]{biblatex}
\DeclareNewSectionCommand[
  style=section,
  level=\subsubsectionnumdepth,
  afterskip=1.5ex plus .2ex,
  beforeskip=-1pt,
  indent=0pt,
  font=\mdseries\small,
  tocstyle=gobble
]{subsecsubtitle}

\newcommand\tvs[1][.5em]{%
  \textcolor{magenta}{
  \makebox[#1]{%
    \kern.07em
    \vrule height.3ex
    \hrulefill
    \vrule height.3ex
    \kern.07em
  }
}}

\newcommand\tvt[1][2em]{%
  \textcolor{magenta}{
  \makebox[#1]{%
    \kern.07em
    \vrule height.3ex
    \hrulefill
    \vrule height.3ex
    \kern.07em
  }
}}

\newcommand\nl{%
  \textcolor{magenta}{\textbackslash{}n}
}

\begin{document}

\newpage

\title{regAL: Python Package for Active Learning of Regression Problems}

\author{Elizaveta Surzhikova$^{1}$, Jonny Proppe$^{1,}$\thanks{\textbf{E-mail}: \url{j.proppe@tu-bs.de} $\vert$ \textbf{Web}: \url{proppe-group.de}}}

\date{\today}

\maketitle

\vspace{-1cm}

\begin{center}
$^1$TU Braunschweig \\ Institute of Physical and Theoretical Chemistry \\ 
Gauss Str 17, 38106 Braunschweig, Germany \\
\end{center}

\begin{abstract}
{\footnotesize
\noindent\textbf{Abstract}\\
Increasingly more research areas rely on machine learning methods to accelerate discovery while saving resources.
Machine learning models, however, usually require large datasets of experimental or computational results, which in certain fields\,---\,such as (bio)chemistry, materials science, or medicine\,---\,are rarely given and often prohibitively expensive to obtain. 
To bypass that obstacle, \textit{active learning} methods are employed to develop machine learning models with a desired performance while requiring the least possible number of computational or experimental results from the domain of application.
For this purpose, the model's knowledge about certain regions of the application domain is estimated to guide the choice of the model's training set.
Although active learning is widely studied for classification problems (discrete outcomes), comparatively few works handle this method for regression problems (continuous outcomes).
In this work, we present our Python package \textit{regAL}, which allows users to evaluate different active learning strategies for regression problems.
With a minimal input of just the dataset in question, but many additional customization and insight options, this package is intended for anyone who aims to perform and understand active learning in their problem-specific scope.\\
}
\end{abstract}

\begin{abstract}
{\footnotesize
\noindent\textbf{Program summary}\\
\textit{Program title:} regAL$^{2}$\footnotetext[2]{\textit{regAL} is an acronym for \textit{active learning for regression}. When we speak German, however, we pronounce it as \textipa{[\textinvscr e"ga\textlengthmark l]} (meaning \say{shelf} in German).}\\
\textit{Program source:}\\ https://git.rz.tu-bs.de/proppe-group/active-learning/regAL\\
\textit{Programming language:} Python 3+\\
\textit{Program dependencies:} numpy, scikit-learn, matplotlib, pandas\\
}
\end{abstract}

\section{Introduction}
Exploring problem spaces is an everyday task in various research fields, often aimed at the prediction or optimization of desired outcomes.
With their increased availability and development, machine learning methods allow promising new approaches for exploring, e.g., chemical compound spaces \cite{von_2020, gryn2024}.
Trained on a suitable dataset, machine learning models can learn from outcomes of known samples and predict outcomes of similar unknown samples efficiently and with less resources required.
While in many research fields \cite{kriz2009, ekel2016} an abundance of training data is available, other fields, such as molecular or materials sciences, often lack sufficiently large datasets of experimental or computational results\,---\,referred to as \textit{labels}\,---\,for training good models, as label generation is very expensive in these areas.
Under such circumstances, machine learning approaches can quickly become unfeasible, despite their strong potentials.
Therefore, in order to achieve a predefined model performance with the least labeling effort, meaning the effort of producing computational or experimental results, \textit{active learning} was introduced \cite{sett2012}.\\
The key idea of active learning is to let an algorithm guide the selection of the model's training set by identifying the most informative samples during runtime, instead of having to provide a fixed training set upfront.
The choice to include an individual training sample is hereby made based on the potential knowledge gain the sample provides to the model.
This gain can be estimated with a variety of different methods.
For example, training samples are commonly picked from areas which are not well sampled yet.
It is important to note, however, that the performance of different active learning methods heavily depends on the dataset to be learned \cite{sett2012}.\\
Although active learning shows promising potential and already finds applications in numerous tasks of various research fields \cite{lang2016, belu2018, tuia2011, erte2007, bowr2004}, it is mostly utilized and researched for classification tasks \cite{sett2012, cai2013}, meaning predictions of \textit{discrete} outcomes.
Regression tasks, on the other hand, are concerned with \textit{continuous} outcomes.
As the assessment of a model's potential knowledge gain is fundamentally different between classification and regression problems, active learning methods for classification either need to be adapted or developed differently to suit regression scopes \cite{one2017, cai2013, frey2014}.
Since active learning is not a universal fix for every problem, but rather has to be partially tailored to the problem at hand, the conceptualization and implementation of necessary algorithms may require resources that are not always available.\\
Although some active learning tools already exist, they focus on a different problem scope.
As already discussed above, due to the previous emphasis on active learning for classification, many works focus entirely on classification tasks \cite{lu2023, tang2019, yang2017}, which are often not transferable to regression problems.
Furthermore, a significant portion of existing works utilize neural networks as target models \cite{holz2023, 2024b, eckh2021}, although Bayesian kernel methods provide a more data-efficient alternative, which is particularly advantegeous for smaller data sets.
Lastly, a majority of developed tools are designed purely for performing active learning, rather than providing custom benchmarking options \cite{novi2021, dank2018}.\\
Here, we introduce our Python package \textit{regAL}.
Our goal was to design a tool for users of all backgrounds to try or perform different active learning approaches specifically for regression problems, to assess whether they are worth investing resources into.
To achieve that, \textit{regAL} can be executed in benchmark mode. 
In this mode, users provide an already labeled dataset and the size of the subset with which the model starts the training in order to benchmark various sample selection methods.
Thereby users can evaluate which methods work best on their particular datasets, freeing them of the burden of performing multiple trial experiments to determine the best method for similar unlabeled datasets.
The entire benchmarking procedure and the model training happen in a black box, showing the user solely the final results, but are still fully transparent.
During any use of our package, a run history is created, which captures all parameters for each sample selection iteration.
Users therefore have access to detailed insights into the model's training procedure, to analyze active learning as a method further beyond just its end result.
In both cases, users can ultimately decide whether active learning is suitable for similar but unlabeled datasets, and\,---\,if so\,---\,can execute \textit{regAL} in learn mode to actually perform active learning on unlabeled data.\\
Although \textit{regAL} can be used with just a dataset as input, the package provides many different customization options.
While, for example, the default trained model is a Gaussian process regressor with an RBF kernel, users can easily change the kernel through a passed argument or overwrite the entire model class to use a specific model architecture.\\
We additionally provide a demonstration of our package on the QM9 dataset \cite{rudd2012, rama2014}, which is a well-established molecular dataset for benchmarking machine-learning methods.
This particular dataset serves a purely exemplary role in the context of this work, as \textit{regAL} is applicable to regression problems in general.

\section{Active Learning}
\begin{figure}[h]
\centering
  \includegraphics[width=8cm]{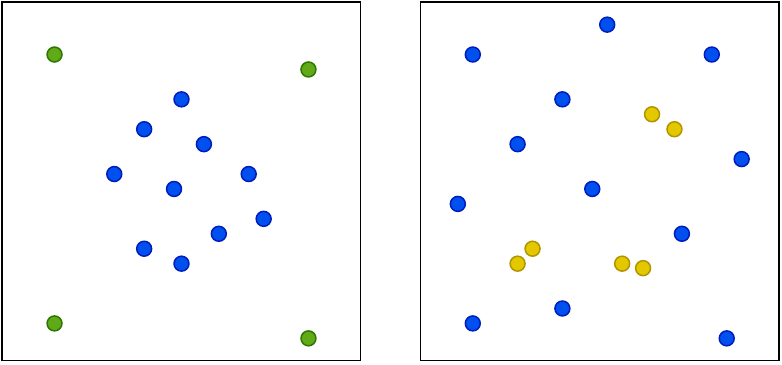}
  \caption{Exemplary spaces with data samples. On the left in green, samples which contribute highly to the model's understanding of the space are shown. On the right, redundant sample pairs, of which the second sample provides negligible additional information to the model after the first sample is labeled, are shown in yellow.
  }
  \label{fig:distribution}
\end{figure}

As machine learning requires datasets of sufficient size, it becomes inaccessible in settings where labeled data is scarce and labeling is expensive.
Therefore, the question arises whether a model with a desired performance can be trained on a subset of the data that would have been needed in the first place.
Regarding a dataset within its domain of application, it quickly becomes apparent that not all samples contribute equally to the model's knowledge gain.
As is illustrated in Fig.~\ref{fig:distribution} (left), it is evident that the green samples contribute highly to the model's understanding of the space, as they cover underrepresented areas.
However, when two samples lay close to each other in the learned space, as shown by the yellow samples in Fig.~\ref{fig:distribution} (right), the additional information of labeling the second sample is insufficient for the additional required resource expenditure.
It is therefore obvious that a model can be trained on a subset of a dataset with negligible performance loss compared to training on the whole dataset.
While this subset can be chosen either randomly or based on the data distribution with approaches such as farthest-first traversal \cite{rose1977}, samples truly important for capturing the underlying domain might be missed.
Samples therefore need to be chosen based on their \textit{informativeness} (how much they reduce the uncertainty of the model's predictions) or their \textit{representativeness} (how well they represent the pattern of the underlying data distribution) \cite{sett2009, huan2014}.
To make sure that only samples which fulfill those criteria are labeled, a branch of machine learning called \textit{active learning} was developed \cite{sett2012}.\\
Active learning is a machine learning method, which iteratively develops a target model with a desired performance with the least possible number of known labels, meaning computational or experimental results.
An active learning procedure starts by training the model on a small initially labeled dataset, to which in each iteration new and previously unknown samples are added based on either their provided informativeness or representativeness (or both) to the model.
When a new unlabeled sample is then chosen, it gets labeled by an oracle\,---\,a hypothetical entity that knows the true label of any sample.
This oracle can be a human expert, a database, an experiment, a simulation or other similar sources.
After being labeled, the sample is added to the model's training set and the model is retrained.\\
Although the core idea of active learning is always the same, implementations vary in the availability of unlabeled data, the sample selection criteria and the size of the selected sample set in each step. 
These variables are discussed in the following subsections.

\subsection{Pool-Based vs Stream-Based Setting}
Depending on the problem setting, unlabeled data may or may not be provided upfront.
In the \textit{pool-based} setting, all of the available, unlabeled samples are known from the beginning.
The whole set of the unlabeled samples is referred to as the \textit{pool}. 
With that, the sample selection algorithm can compare individual samples with all other available samples and choose the best fitting one to be labeled next, based on a given criterion \cite{stol2024, sett2012, kuma2020}.
This approach has the advantage that the sample selection algorithm can with certainty make the best informed decision.
However, if the pool contains a large number of samples, calculations can become very expensive and corresponding approaches rather slow (although in most cases still significantly faster than labeling).\\
If the whole pool is unknown at the beginning of the experiment, but new unlabeled samples are being added continuously, the \textit{stream-based} setting applies.
Here, the learning model is presented with one individual sample (or a small set of samples) at a time and needs to make an instant decision to keep or discard that sample \cite{sett2012, kuma2020, simm2018, smit2018, herb2023}.
Compared to the pool-based setting, this approach is faster and requires less memory with increasing dataset size.
The learning model's decisions, however, are less informed during early iterations, as it has no knowledge on the following data stream and not enough already chosen samples for comparison.\\
A last method to collect data for the learning model's training set is to create training samples from scratch, called the \textit{synthetic} approach \cite{wang2015, sett2012, kuma2020}.
This method allows to generate an arbitrary amount of samples and therefore compose the training set from an essentially \say{infinite pool}.
Thus, after learning the problem scope, the synthesis algorithm generates samples, which optimally cover the entire problem domain, while using as little instances as possible.
While this approach is favorable if the provided, unlabeled data is of low quality or does not cover the entire problem domain, the generated samples may be non-existent (e.g., non-equilibrium, high-energy chemical structures) and therefore induce incorrect information into the training set.
Without additional control mechanisms this approach is therefore not suitable for settings in which the data distribution is discrete, such as the chemical compound space.\\

\noindent In the current version of \textit{regAL}, only the pool-based setting is considered.

\begin{figure*}[h]
\centering
  \includegraphics[width=\textwidth]{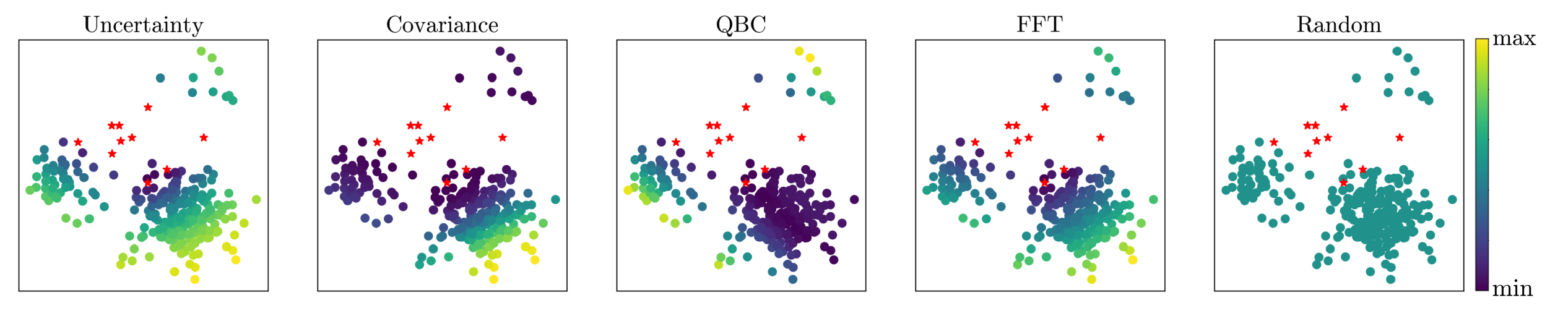}
  \caption{Representation of all sample selection methods currently implemented in \textit{regAL}. The red stars constitute the training set of the model, while the dots are the unknown pool. The color of the dots indicates their priority to be labeled, with yellow and blue representing the highest and lowest priorities, respectively. Note that this ranking is only valid for this particular iteration and will change after the model is retrained.}
  \label{fig:contour}
\end{figure*}

\subsection{Sample Selection}
\label{sec:sample_selection}
In the pool-based setting, the choice on the next training sample is made based on its provided information gain to the learning model.
Depending on the respective definition of \say{information gain} and the requirements for the specific model, this can be realised with different methods.
In the following we will distinguish between \textit{model-free} and \textit{model-based}, as well as \textit{active} and \textit{passive} methods.
Methods that require information from the trained model, such as its uncertainties about unlabeled samples, are \textit{model-based}.
On the other hand, methods that operate without taking the model into account are \textit{model-free}.
Considering from a different perspective, \textit{active} methods are influenced by the current state of experiment components, such as the model output or the current training set, to make decisions about new samples.
Inversely, \textit{passive} methods do not take the progression of the experiment into account.

\subsubsection*{Uncertainty Sampling}
\subsecsubtitle*{\textit{model-based, active}}
\label{uncert}
Perhaps the most intuitive sample selection method is \textit{uncertainty sampling}.
Here, the learning model $\theta$ predicts the label $y$ of each unlabeled sample $\mathbf{x} \in {\cal U}$, along with the uncertainty $U_\theta(\hat{y}|\mathbf{x})$ about that label.
Here, $\hat{y}$ corresponds to the prediction of $y$.
The higher the model's uncertainty about a label, the less knowledge the model has in that sample's area, as visualized in Fig.~\ref{fig:contour}.
The model would therefore select the sample with the highest uncertainty $U_\theta$, 
\begin{equation}
\label{eq:unc}
    \mathbf{x}^* = \arg \max_\mathbf{x} U_\theta(\hat{y}|\mathbf{x}) \ \forall \mathbf{x} \in {\cal U} \ ,
\end{equation}
as labeling this sample would give the model the most amount of new information about the problem domain \cite{sett2012}.\\
Although this method comes at a low cost for models where a prediction uncertainty is given along with the prediction value, such as Gaussian process regressors, it is difficult to implement for models where this is not the case.
In the case of neural networks, various approaches are known to provide the uncertainty along with a prediction \cite{hirs2020, jane2019, zhou2023}.

\subsubsection*{Covariance Sampling}
\subsecsubtitle*{\textit{model-based, active}}
Similar to uncertainty sampling, \textit{covariance sampling} incorporates the uncertainty $U_\theta$ of the unlabeled samples $\mathbf{x} \in {\cal U}$.
Additionally, however, the sample's pairwise covariances with all other samples in ${\cal U}$ are taken into account,
\begin{equation}
\label{eq:cov}
    \mathbf{x}^* = \arg \max_\mathbf{x} U_\theta(\hat{y}|\mathbf{x}) \sum_{\substack{\bar{\mathbf{x}} \in {\cal U}\\\bar{\mathbf{x}} \neq \mathbf{x}}}\text{cov}(\mathbf{x}, \bar{\mathbf{x}}) \ \forall \mathbf{x} \in {\cal U}
\end{equation}
with
\begin{equation}
\label{eq:cov2}
    \text{cov}(\mathbf{x}_i, \mathbf{x}_j) = \frac{1}{N-1} \sum_{k=1}^{N}(\mathbf{x}_{i_k}-\mu_{\mathbf{x}_i})(\mathbf{x}_{j_k}-\mu_{\mathbf{x}_j}) \ .
\end{equation}
The sample choice based on those criteria can be easily recognized in Fig.~\ref{fig:contour}.
Therefore, the unlabeled pool sample with the highest expected variation individually or in relation to other samples is chosen.\\
As discussed with uncertainty sampling, this method is inapplicable in settings where a model's prediction uncertainty is unknown.
Furthermore, this method is significantly more expensive compared to uncertainty sampling, as it involves operations on the entire covariance matrix, unlike just its main diagonal.
An exemplary runtime comparison follows in Sec.~\ref{runtime}.

\subsubsection*{Query by Committee (QBC)}
\subsecsubtitle*{\textit{model-based, active}}
A different approach to quantifying the learning model's uncertainty is through the \textit{query-by-committee} (QBC) method \cite{seun1992}.
Traditionally, this method is performed by training a committee of neural networks with different architectures on the same training set.
The unlabeled samples $\mathbf{x} \in {\cal U}$ in question are then predicted by each model $\theta$ in the committee ${\cal C = \{\theta\}}$, and the sample for which the predictions $P$ of the committee diverge the most is chosen as the next sample to be labeled  \cite{yang2024a, smit2019, stol2024, behl2021},
\begin{multline}
\label{eq:qbc}
    \mathbf{x}^* = \arg \max_\mathbf{x} \big[\max(P_\theta(\hat{y}|\mathbf{x})) - \min(P_\theta(\hat{y}|\mathbf{x}))\big]\\ \forall \mathbf{x} \in {\cal U}, \ \forall \theta \in {\cal C} \ .
\end{multline}
An example of this method can be seen in Fig.~\ref{fig:contour}.
When working with kernel-based models, to which Gaussian processes belong, the model architecture can not be altered in the same way as with neural networks, where the number of layers and neurons can be freely adjusted to modify the model's complexity.
The committee ${\cal C}$ is therefore trained on varying subsets of the training set or with different kernels.
In this work, we chose the latter.\\

\noindent To be able to compare the performance of the above mentioned sample selection strategies, \textit{regAL} also performs sample selection with model-free methods, both active and passive.

\subsubsection*{Random Sampling}
\subsecsubtitle*{\textit{model-free, passive}}
The baseline sample selection method is \textit{random sampling}. As the name suggests, this method picks samples from the unlabeled pool ${\cal U}$ at random without any specific strategy,
\begin{equation}
    \mathbf{x}^* = \text{rand}(\mathbf{x}) \ \forall \mathbf{x} \in {\cal U} \ .
\end{equation}
This is shown in Fig.~\ref{fig:contour}, as all unknown samples have the same color, meaning their chances of being labeled next are all equal.
If an active learning sample selection method is not suitable for a given dataset, its results will be closer to those of random sampling or even worse.

\subsubsection*{Farthest-First Traversal (FFT)}
\subsecsubtitle*{\textit{model-free, active}}
A different model-free sample selection method is \textit{farthest-first traversal} (alt.\ called \textit{kernel-farthest-first}) \cite{rose1977}.
Starting from a random initial sample, each new to-be-labeled sample $\mathbf{x} \in {\cal U}$ is chosen to be the one farthest away from its closest labeled neighbour $\bar{\mathbf{x}} \in {\cal L}$,
\begin{equation}
\label{eq:fft}
    \mathbf{x}^* = \arg \max_\mathbf{x} \big[\min(d(\mathbf{x}, \bar{\mathbf{x}})) \big] \ \forall \mathbf{x} \in {\cal U}, \ \forall \bar{\mathbf{x}} \in {\cal L} \ ,
\end{equation}
according to their Euclidean distance,
\begin{equation}
\label{eq:dist}
    d(\mathbf{x}_i,\mathbf{x}_j) = \sqrt{\sum_{i=1}^{N}(x_i - x_j)^2} \ .
\end{equation}

In the example of Fig.~\ref{fig:contour}, this predominantly applies to the unknown samples at the edges of the space, as all labeled samples lie in the middle.
As this approach is based solely on the distribution of the data pool in the space and does not rely on the model's learning process, it is categorized as \textit{model-free}.
However, as the currently labeled samples determine the position of the next to-be-labeled sample in each iteration, this approach can be considered \textit{active}.
It is therefore often successfully used to perform active learning and is implemented in \textit{regAL} as a benchmark for the model-based active learning approaches mentioned above.\\

\noindent While the approaches above were chosen for the current implementation of \textit{regAL}, it is by no means a complete collection of all possible sample selection strategies.
A variety of other methods with different approaches, such as maximizing expected model change \cite{cai2013}, density and diversity \cite{one2017}, or expected model output change \cite{frey2014} are used to perform active learning.
Although these and further sample selection methods, including stream-based sampling, are currently not available in \textit{regAL}, their integration is under consideration.

\subsection{Singular vs Batch-Wise Sampling}
\label{sec:sing_bw}
Classically, active learning is performed by choosing and labeling one new training sample per iteration.
As in many settings the capacity for multiple parallel calculations or experiments is available, active learning is increasingly performed in a batch-wise setting, which significantly saves time.\\
In contrast to singular sample selection, in every iteration multiple samples forming a batch are chosen and labeled, instead of a single sample \cite{hoi2006}.
This approach, however, also requires different sample selection methods \cite{prop2019a, kirs2019, ash2021, holz2023, zave2022}, as the methods described in Sec.~\ref{sec:sample_selection} require updated model parameters after each chosen sample, to make a considerate decision on the next sample.
Otherwise, these sample selection methods would predominantly choose similar samples with close proximity, as samples in the same region of the target space likely show similar characteristics from the model's perspective.
We currently limit the scope of our package to single sample selection methods, but consider its extension to batch-wise selection methods in the future.

\subsection{Initial set size}
In order for a machine-learning model to make considered decisions on its areas of uncertainty, it has to pass an initial training cycle.
Therefore, a small subset of the data pool has to be labeled prior to starting the active learning experiment.\\
The size of this initial dataset can not generally be defined for all cases, but rather depends on the data distribution at hand.
While for some datasets, where the structure of the underlying distribution is easy to grasp for the learning model, only few initial samples are sufficient, other datasets may require a substantial amount of samples to properly train the model.
It is therefore important to choose the size of the initial training set for each dataset carefully, so that proper training of the model is ensured, while not wasting labeling effort on unnecessary samples.
\label{sec:init}

\section{Software}
\label{sec:mode}
\begin{figure*}[h]
  \includegraphics[width=\textwidth]{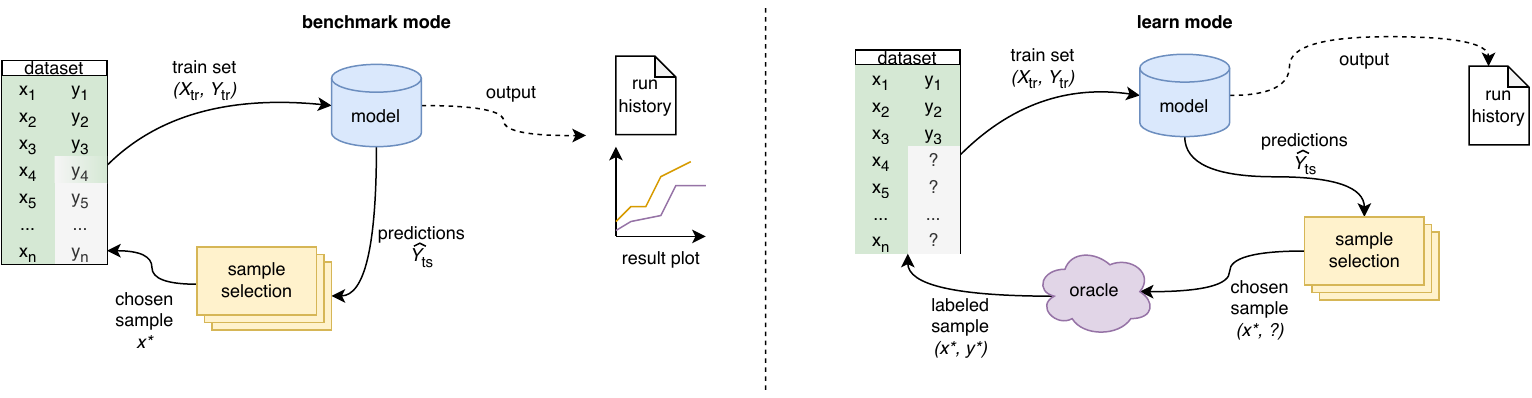}
  \caption{Illustration of benchmark and learn mode procedures in \textit{regAL}. In benchmark mode, the labels of all samples are known from the start, but are only incrementally revealed to the model. In learn mode, labels are known only for some samples, and new labels are generated by an oracle, which can be user inputs, rule-based decisions, results of experiments or computer simulations, etc.}
  \label{fig:flowchart}
\end{figure*}

Our package is entirely written in Python, as it is one of the most commonly used programming language in scientific environments and provides well-equipped machine learning libraries.
We base the entire functionality of our package on the \textit{scikit-learn} package, since it is widely popular and often times already installed on user devices.
In order to offer users an obstacle-free introduction to using active learning, our package is intended to be used as a black box.
Nevertheless, all relevant experiment parameters are tracked during runtime and can be accessed after the experiment for further analysis, if needed.\\
To suit a wide spectrum of user requirements, our package can be used for two different purposes.
For users who are not familiar with active learning entirely, who want to try new sample selection methods, or evaluate the performance of active learning on custom datasets, we implemented a \textit{benchmark mode}.
Here, an already labeled dataset is passed to the algorithm, which pretends to not know the labels, but uses them for \say{labeling} the selected training samples, as shown in Fig.~\ref{fig:flowchart} (left).
The aim of this mode is to provide users with a demonstration of how different active learning method behave on their specific data.
Users can then transfer this knowledge on their similar but unlabeled datasets, thereby saving the resources of failed active learning attempts on the unlabeled data.\\
To then actually perform active learning on unlabeled data, the \textit{learn mode} can be used.
In this mode, the data is provided to the algorithm without the labels, however with an additional function through which the labels can be obtained, as shown in Fig.~\ref{fig:flowchart} (right).
This function is provided by the user and can therefore be implemented to call any other function, start another program or automated experiment, or prompt the user for the label of a training sample.
Regarding automated experimentation, \textit{regAL} can also be integrated into self-driving laboratories \cite{abol2023}.

\subsection{Performance Example on the QM9 Dataset}
In this section we show the features and performance of the \textit{regAL} package on a uniformly sampled subset of the QM9 dataset \cite{rudd2012, rama2014}.
QM9 is a dataset containing quantum chemical results for about 134k small organic molecules with up to nine non-hydrogen atoms (carbon, nitrogen, oxygen, and fluorine).
For each of those molecules the dataset provides equilibrium structures and 15 different properties, such as the dipole moment, molecular orbital energies, and rotational constants.
We chose this dataset as it is rather large, widely known and often used for benchmarking machine-learning models in chemical contexts \cite{unke2018, gilm2017, hou2018}.

\subsubsection*{Experiment Setup}
\begin{figure}[h]
\centering
  \includegraphics[width=8.5cm]{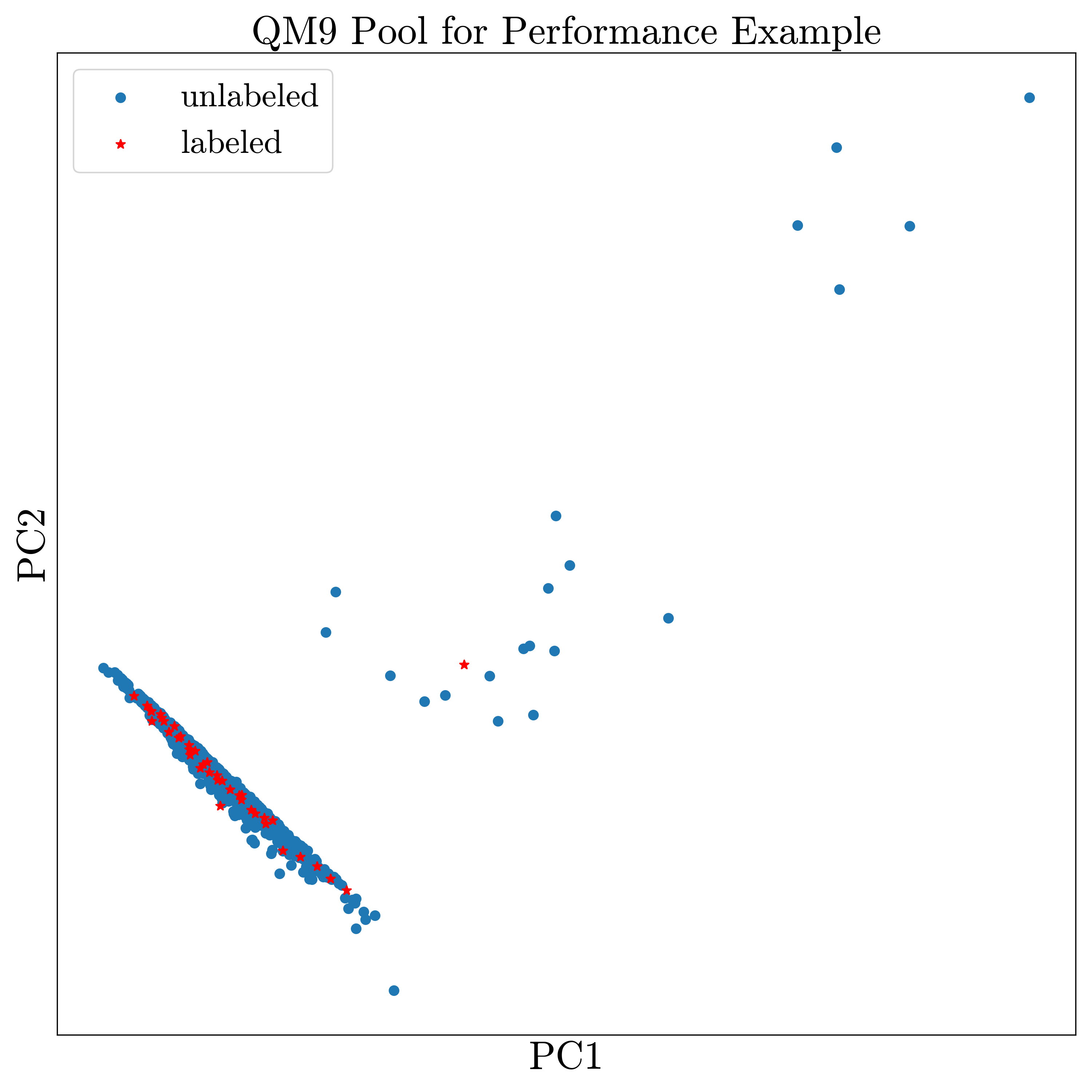}
  \caption{Visualization of the QM9 subset used for the performance example (see also Fig.~\ref{fig:output_plot}). 1340 points were uniformly sampled from the QM9 dataset, out of which 40 were randomly chosen to be in the initial training set, shown by red stars.
  }
  \label{fig:qm9_dataset}
\end{figure}

For this example experiment we predicted the HOMO energies ($E_\text{HOMO}$) contained in the QM9 database from the invariant two-body interactions descriptor $F_\text{2B}$ \cite{pron2018}. 
For the to-be-trained model we used the default Gaussian process regressor with a scaled RBF kernel,\\
\begin{equation}
    k_\text{RBF}(\mathbf{x}_i,\mathbf{x}_j) = \sigma_\text{f}^2 \exp\left(-\frac{d(\mathbf{x}_i,\mathbf{x}_j)^2}{2l^2} \right) \ ,
\end{equation}
with $d(\mathbf{x}_i,\mathbf{x}_j)$ defined in Eq.~(\ref{eq:dist}).
The hyperparameters $l$ (length scale) and $\sigma_\text{f}^2$ (functional variance) are subject to optimization.
In \textit{scikit-learn}, this kernel can be realized via:
\begin{center}
    \texttt{ConstantKernel() * RBF()}
\end{center}
The initial training dataset consisted of 40 randomly selected samples from the pool, the experiment was performed for 100 iterations, and all available sample selection methods were tested.
A visualization of the whole pool along with the 40 initial training samples is shown in Fig.~\ref{fig:qm9_dataset}.
As due to being represented by the $F_\text{2B}$ descriptor each sample had 225 dimensions, a principle-component analysis was performed, and the first two principle components are shown.

\subsubsection*{Experiment Results}
\begin{figure}[h]
\centering
  \includegraphics[width=9cm]{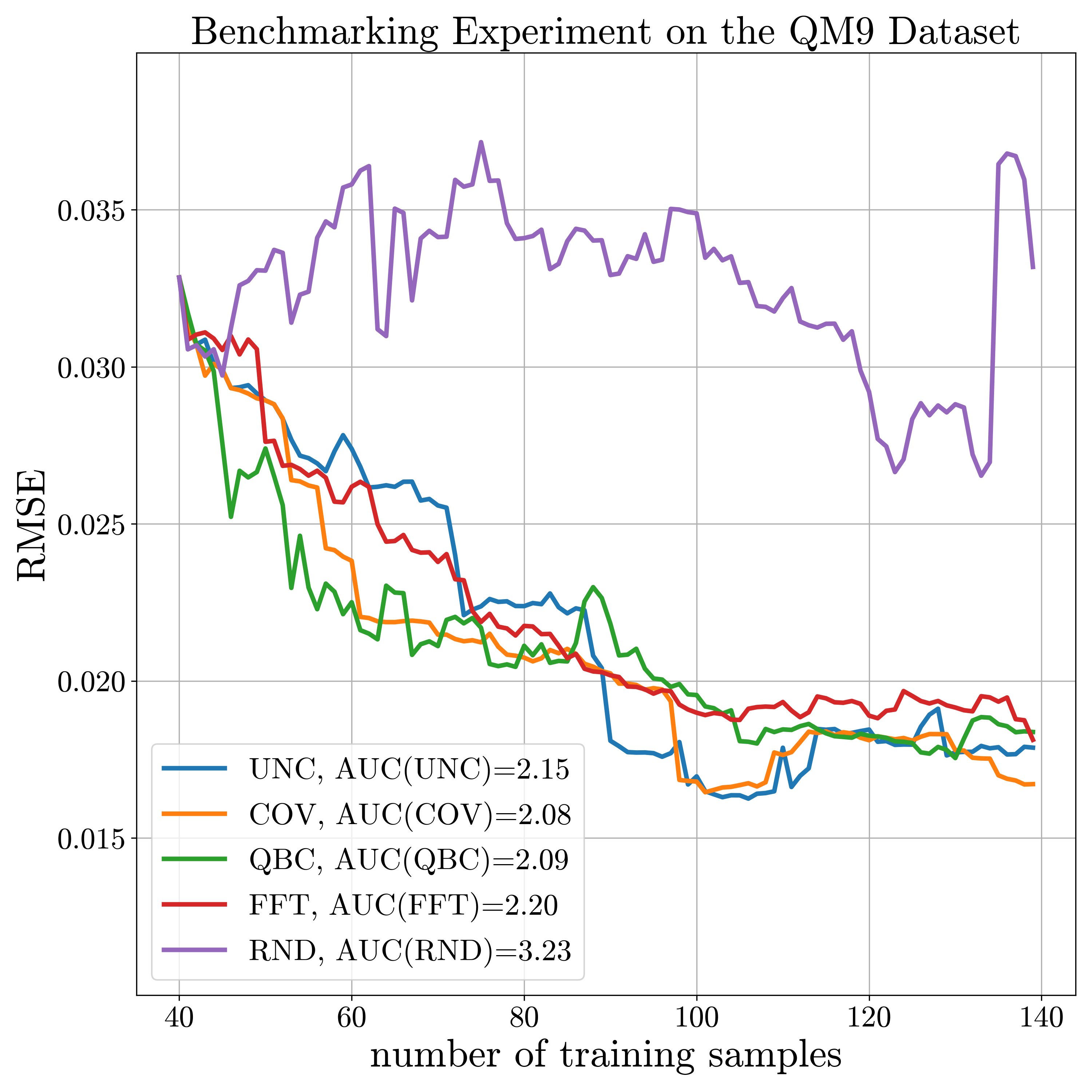}
  \caption{Examplary \textit{regAL} output in benchmark mode, showing the progression of the root mean squared error (RMSE) over the number of active learning cycles for a subset of the QM9 dataset (target: HOMO energy ($E_\text{HOMO}$), features: invariant two-body interactions descriptor $F_\text{2B}$). We abbreviated uncertainty sampling with UNC, covariance sampling with COV and random sampling with RND.
  }
  \label{fig:output_plot}
\end{figure}

The visual output of the benchmarking experiment is shown in Fig.~\ref{fig:output_plot}, which serves as a first impression to the user.
It shows the progression of the root mean squared error (RMSE) of the different sample selection methods over the experiments iterations, in this case from 40 to 140 training samples.
In the plot's legend, the area under the curve (AUC) after the final iteration is listed for each method.
We emphasize that the comparison of RMSE values at a specific training set size is not a suitable performance comparison between different sample selection methods, as these values are fluctuating on a smaller scale.
The crucial indicator for the evaluation of a methods performance is its AUC, as it shows the overall progression of that method's performance.
While different methods may achieve similar RMSE values after the final iteration, some methods achieve lower values, and thereby a better performance, faster than others, which is represented in the AUC.
Additionally, the singular RMSE values are highly sensitive to minor setting changes, such as different randomization seeds for choosing the initial training set.
The AUC however, despite also fluctuating with changing settings, has a consistent and reproducible trend within an experiment.
The choice on the number of active learning iterations should therefore be made based on the available resources and the used sample selection method based on the best AUC for that number of iterations.\\
In this particular example, covariance sampling (COV) and QBC achieve the best results.
As covariance sampling captures informativeness as well as representativeness of samples, it can make highly considerate decisions about the most fitting new samples.
The performance of the QBC method, on the other hand, is typically difficult to interpret, as it is not obvious which committee member gives which prediction for specific samples.
Uncertainty sampling (UNC) and FFT perform very similarly, as in this case they target very similar unlabeled points.
As most of the chosen labeled points lie along the center of the sample ridge on the bottom left, as shown in Fig.~\ref{fig:qm9_dataset}, uncertainty sampling will naturally pick samples at the borders of the ridge and in the separated sample groups, as these areas are most unknown to the model.
Similarly, FFT will choose the same samples, as they are also farthest away from the labeled samples.
The model performance under random sampling (RND) shows very inconsistent behaviour and does not significantly improve over the course of the experiment, as it chooses samples without a specific strategy.
The partial worsening of the model performance for the random-sampling method can be attributed to the trained model getting caught in local optima with respect to its hyperparameters.
We emphasize that those results are highly dependent on the particular dataset and are not transferable to other scopes.\\

\noindent Details of the active-learning procedure in both modes can be found in the respective run history files, as elucidated in the following Sec.~\ref{sec:run_hist}.

\begin{figure}[h]
  \includegraphics[width=9cm]{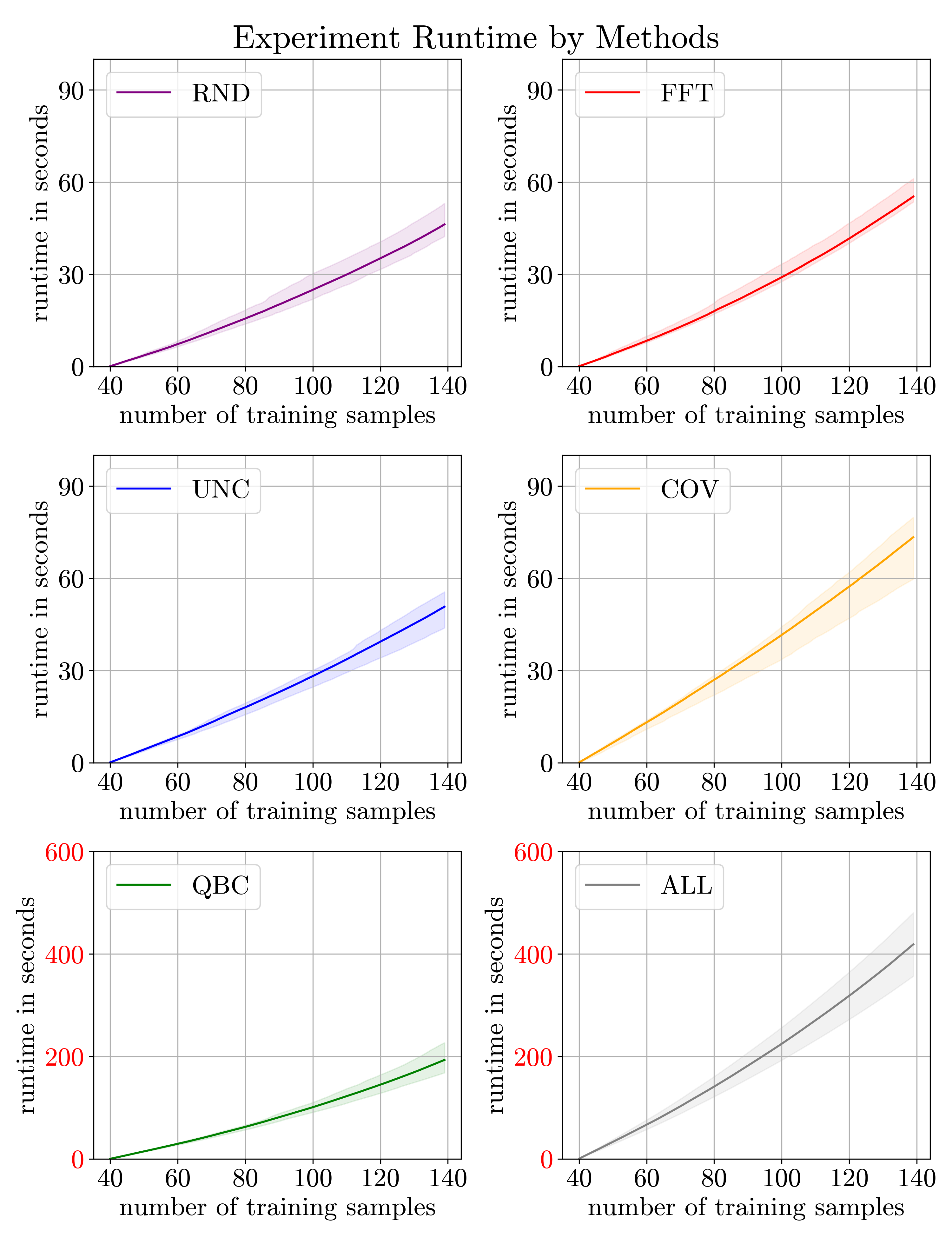}
  \caption{Runtime in seconds versus size of training set for each sample selection method currently implemented in \textit{regAL}, each averaged over ten executions. The bands along each curve show the range between the minimum and maximum runtime values among the ten executions. We abbreviated uncertainty sampling with UNC, covariance sampling with COV and random sampling with RND. ALL shows the summed total runtime of all methods.
  }
  \label{fig:runtime_plot}
\end{figure}

\subsubsection*{Runtime Assessment}
\label{runtime}
As active learning is mostly intended for and performed in settings where labeling is the most time-consuming part of the training procedure, the runtime for the rest of the process should be negligible, to not further extend the whole experiment.
In this section we want to show that this is the case for \textit{regAL} by recording its runtime in benchmark mode.\\
The runtime of an experiment iteration is highly dependent on the chosen sample selection method, as well as the current size of the training set.
This particular assessment was made based on training a Gaussian process regressor with a scaled RBF kernel in benchmark mode on an 8-core M2 CPU and averaged over ten executions using different randomization seeds.
As these plots were recorded in benchmark mode, those values show nearly the pure runtime of the package, as the time for an array readout (to get the new sample labels) is negligible on those time frames.
In learn mode, the runtime will be different depending on the provided oracle function, which is assumed to be the bottleneck of the active-learning loop.
As shown in Fig.~\ref{fig:runtime_plot}, the runtime increases for all methods with the amount of training samples, as model training naturally decelerates with a rising training set size.
The runtime for random sample selection is lowest, as it does not require any further computation for the choice of the next labeled sample.
The FFT, uncertainty (UNC), and covariance (COV) methods all show a similar runtime, with covariance sampling being the slowest, as it involves approximately the square amount of matrix operations compared to uncertainty sampling, for example.
This runtime gap between the covariance and the FFT/uncertainty methods will further increase with larger training set sizes.
QBC, however, stands out significantly over the other methods.
The reason for this behavior is the training of multiple models (in this case three) to form the committee in each iteration.
The runtime curve of the QBC method therefore has a steeper slope compared to the other methods.\\
The runtimes discussed in this section refer to active learning in a single-sampling setting, as described in Sec.~\ref{sec:sing_bw}.
In a batch-wise sampling setting, as also discussed in Sec.~\ref{sec:sing_bw}, the runtime would be significantly lower, as multiple new samples would be added to the training set in each iteration.
Therefore the learning model would only need to be retrained once per batch and not once per every sample.

\begin{figure*}[h]
    \ttfamily
    \begin{framed}
    \#start\tvs{}time:\tvs{}\textcolor{teal}{DDMMYYYY-HHmmSS},\tvs{}mode:\tvs{}\textcolor{teal}{benchmark},\tvs{}sample\tvs{}selection\tvs{}method:\tvs{}\textcolor{teal}{qbc},\tvs{}seed:\tvs{}\textcolor{teal}{5}\textcolor{magenta}{\nl{}}\\
    \#models\textcolor{magenta}{\textbackslash{}t}labeled\_samples\textcolor{magenta}{\textbackslash{}t}labels\textcolor{magenta}{\textbackslash{}t}hyperparams\textcolor{magenta}{\textbackslash{}t}\textcolor{teal}{R2/RMSE}\textcolor{magenta}{\textbackslash{}t}AUC\textcolor{magenta}{\textbackslash{}t}runtime\textcolor{magenta}{\textbackslash{}n}
    \end{framed}
    \caption{Header lines of every run history file produced by \textit{regAL}. Text in black explicitly occurs in the file, turquoise text varies depending on the experiment, and magenta symbols are text separators not visible in the files. Short separators (\tvs{}) represent white spaces. In learn mode the R2/RMSE and AUC keywords are omitted.}
    \label{fig:header}

    \begin{framed}
    \textcolor{teal}{model\_file.pickle\textcolor{magenta}{\textbackslash{}t}[1, 2, 3]\textcolor{magenta}{\textbackslash{}t}[0.5, 0.3, 0.4]\textcolor{magenta}{\textbackslash{}t}\{noise\_level=1e-05, ...\}\textcolor{magenta}{\textbackslash{}t}0.02\textcolor{magenta}{\textbackslash{}t}0.435\textcolor{magenta}{\textbackslash{}t}400\textcolor{magenta}{\textbackslash{}n}}
    \end{framed}
    \caption{Example of a body line in run history files produced by \textit{regAL}. All turquoise text is exemplary, as the items are dependent on particular experiments. Magenta symbols are text separators and not visible in the files. In learn mode the R2/RMSE and AUC entries are omitted.}
    \label{fig:body}
\end{figure*}

\subsection{Additional Features}
In addition to the different operation modes, as described in the beginning of Section~\ref{sec:mode}, our package offers two further features, which are presented in the following.

\subsubsection*{Run History}
\label{sec:run_hist}
\begin{figure}[!ht]
    \centering
    \includegraphics[width=9cm]{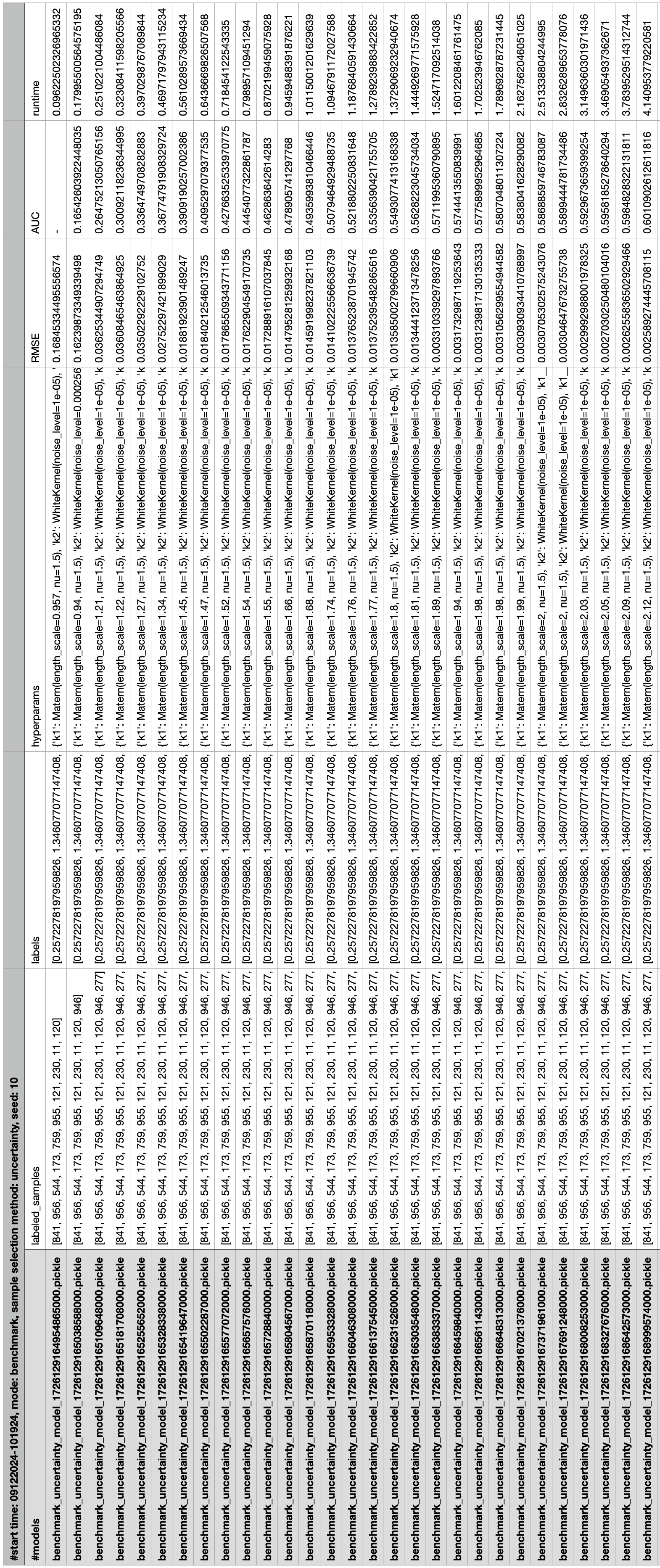}
    \caption{Example run history file produced by \textit{regAL}. Note that the run history files are plain text files and the file in the figure is only opened in a spreadsheet editor for a better structural overview.}
    \label{fig:output_file}
\end{figure}

As this package is intended to not only perform active learning, but also evaluate its performance on a desired dataset, each experiment will document various parameters to a plain text file, which is saved in the user's project folder.
Additionally, in every iteration the current model is saved to a pickle file.\\
The run history file captures all relevant parameters that define the experiment's scope, thereby ensuring its reproducibility, and in each learning iteration tracks its current state.
Parameters that are constant throughout an experiment, namely the benchmark or learn mode indicator (as in the beginning of Section~\ref{sec:mode}), the sample selection method (as in Section~\ref{sec:sample_selection}), and the randomization seed, are added once to the header of the file, as shown in Fig.~\ref{fig:header}.
An example of each following line is shown in Fig.~\ref{fig:body}.
Both examples show the lines in benchmark mode.
In learn mode, the $R^2$/RMSE and AUC columns are omitted.\\
In both figures, all text in black explicitly occurs in the file, while the text in turquoise is exemplary and differs in each file depending on the experiment.
In magenta, the text separators and line breaks are shown.
These are invisible in the file, but are relevant when reading from the file.
Please note that this file will likely look rather unstructured in many text editors, as it is particularly designed to be read and disassembled by a program, and not meant to be primarily human-readable.\\
Using this information, the learning process of the active learning experiment can be tracked and analyzed and thereby studied further beyond just evaluating its end result.\\
A visual example of such a file is provided in Fig.~\ref{fig:output_file}.
Note that the run history files are plain text files and the file in the figure is only opened in a spreadsheet editor for a better structural overview.
Here, the line structures shown in Figs.~\ref{fig:header} and \ref{fig:body} can be seen with actual experimental data.
Each additional line extends the contents of the previous line, so for entries that contain lists, like the labeled samples and the labels, the whole current list is written in each line, as opposed to just the newly added sample.
Similarly for the $R^2$/RMSE, AUC, and runtime columns, the values in each line are accumulated to, e.g., indicate the whole AUC up to that point.
This allows to easily assess the performance of the model or continue the experiment at a specific point by just processing one line from the file instead of additionally processing all preceding ones.
Only the model's hyperparameter column is newly written in every iteration, as it solely refers to the latest model version at that point of the experiment.

\subsubsection*{Resume Training After Abort}
\begin{figure*}[h]
  \includegraphics[width=\textwidth]{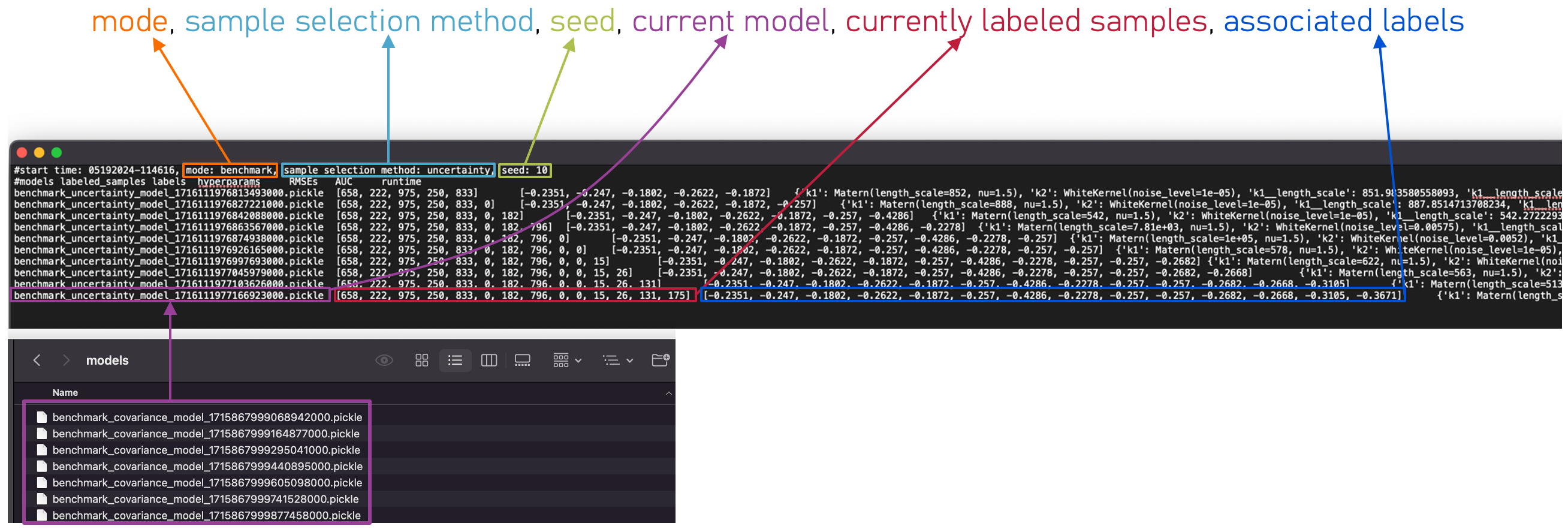}
  \caption{Recomposition of the \textit{regAL} experiment parameters from an example run history file. 
  This is used to resume an active learning experiment after it has ended.}
  \label{fig:resume}
\end{figure*}

In many scenarios, model training can take a substantial amount of time to finish.
In other cases users may want to perform additional training cycles after their active learning experiment is completed, without having to rerun it from the beginning.
For such situations our package provides the \textit{resume-after-abort} functionality.
This functionality relies on the documentation of the run history and the saved model files, as described in the subsection above.
Among others, this function takes the run history file as input, from the first and last line of which it derives the experiment settings and the latest experiment variables, such as the current model filename, the currently known samples and their labels, as is shown in Fig.~\ref{fig:resume}.
It then loads the respective model file, reinitializes the experiment and resumes it from its previous end.
To ensure consistency between the start and the continued experiment, the fixed experiment settings, such as the sample selection method or the randomization seed, cannot be changed.
The experiment can then be continued starting from the last point of the started experiment for a chosen amount of additional iterations.

\subsection{How To Use}
\subsubsection*{Prepare AL Experiment}
A short example code on how to run a dummy experiment is shown in Figs.~\ref{fig:example_code_1}--\ref{fig:example_code_3}.
The minimal necessary imports are listed in lines 1--3 of Fig.~\ref{fig:example_code_1}.
In the \texttt{get\_random\_data()} function in lines 6--10 of Fig.~\ref{fig:example_code_1}, a dummy dataset is defined.
This is just for demonstration purposes and should be replaced by a custom function to get the desired data, which will look different in each individual experiment.
In line 13 of Fig.~\ref{fig:example_code_1}, a model object from the \texttt{ALModel}-class is created.
The model class is prescribed with a default model architecture and functions to fit, make predictions with and obtain the prediction uncertainties from the model.
We chose the default model to be \textit{scikit-learn}'s Gaussian process regressor, as it performs well on small datasets and provides the prediction uncertainty along with the predictions for free.
When using the default model architecture, users can select one of the prescribed kernels.
A custom model architecture can also be provided.
In this case, the user is responsible to overwrite the \texttt{ALModel}-class and all of its functions to fit the desired model architecture.
In line 15 of Fig.~\ref{fig:example_code_1}, the data and labels are loaded from the previously defined data acquisition function.
The following procedure now differs for benchmark and learn mode.

\subsubsection*{Run Benchmarking Experiment}
To use the package in benchmark mode, see the example shown in Fig.~\ref{fig:example_code_2}.
Using the model object, the training data and its labels, an experiment object from the \texttt{ALExperiment}-class is created in line 18.
This is the base for the active learning experiment.
We chose to define each experiment as an individual object, as it allows to run multiple experiments with different parameters and compare their outcomes, without them being overwritten.
The variable \texttt{experiment1} now contains all information corresponding to this particular experiment setting.
After creating a class object with the desired settings, the experiment's \texttt{perform\_al()}-function is called in line 20 with the parameter \textsf{benchmark}.
This function takes the desired size of the initial \say{known} training set and a list of the to-be-tested selection methods as parameters.
\textit{regAL} then performs active learning benchmark experiments on the previously defined model, data and labels for the active learning sample selection methods passed.
When performing QBC, a custom committee of models can also be passed.
If none is provided, a committee of Gaussian process regressors with three different kernels (\texttt{Matern(nu=0.5)}, \texttt{Matern(nu=1.5)} and \texttt{RBF()}\,---\,each multiplied with the \texttt{ConstantKernel()}) is trained.

\subsubsection*{Run Learning Experiment}
To perform learning instead of benchmarking, refer to Fig.~\ref{fig:example_code_3}.
Here, an experiment object from the \texttt{ALExperiment}-class is again created in line 21, except this time leaving out the labels parameter, as the labels are not known yet.
An array for the labels is created internally when the experiment object is instantiated and is filled during the learning procedure.
In line 23, the function \texttt{perform\_al()} is called with the \textsf{learn} parameter, the number of iterations and a list of the selection methods to be performed.
In contrast to the benchmark example, no initial set size is passed and instead the indices of the initially known data points from the \textit{data} array and an array with the corresponding labels are passed.
Lastly, an oracle function needs to be provided.
In line 25, a placeholder oracle function is shown.
This function does not need to be a simple in-place calculation as shown in the example, but can also be a readout from a file, a start of a different experiment, or a user input request.

\subsubsection*{Resume Experiment}
If a benchmark or learn experiment is interrupted and needs to be resumed or needs to be extended by more iterations, the \texttt{resume\_al()}-function can be called as shown in line 21 of Fig.~\ref{fig:example_code_2} for a benchmark experiment and line 25 of Fig.~\ref{fig:example_code_3} for a learn experiment.
This functions takes the name of the run history file of the corresponding experiment and the number of additional iterations as input.
For a learning experiment, the oracle function needs to be passed as an additional input.
In case of the QBC method, a custom model committee can again also be passed as input.\\

\noindent
Not all function input parameters and customization options were shown in this example.
For more details on all function inputs, please consult the project description.$^{3}$\footnotetext[3]{https://git.rz.tu-bs.de/proppe-group/active-learning/regAL/-/blob/main/README.md}

\begin{figure}[h]
\begin{lstlisting}[language=Python]
import numpy as np
from regal import ALExperiment
from regal.model import ALModel

# create random data
def get_random_data():
     samples = np.random.random((1000, 1))
     labs = samples ** 2

    return samples, labs

# use default model
model = ALModel("rbf")
# get data and labels from function above
data, labels = get_random_data()
\end{lstlisting}
\caption{Code example on how to prepare an active learning experiment in \textit{regAL}.}
\label{fig:example_code_1}

\begin{lstlisting}[language=Python]
# create object for model, data and labels
experiment1 = ALExperiment(model, data,
               labels=labels, seed=0)
# benchmark dataset for 20 iterations
experiment1.perform_al("benchmark", 20, 
               selection_methods=["qbc", "uncertainty"],
               init_set_size=10, error="rmse")
# resume benchmarking for 5 additional iterations
experiment1.resume_al("output_benchmark_qbc.txt", 5)
\end{lstlisting}
\caption{Code example on how to start \textit{regAL} in benchmark mode.}
\label{fig:example_code_2}

\begin{lstlisting}[language=Python]
# define labeling function
def oracle(x):
    return x**2

# create object for model, data and labels
experiment2 = ALExperiment(model, data, seed=0)
# learn dataset with 3 known samples
experiment2.perform_al("learn", 20, 
               selection_methods=["qbc", "uncertainty"],
               known_data=data[[2, 20, 60]], 
               known_labels=np.array([....]),
               func=oracle)
# resume learning for 5 additional iterations
experiment2.resume_al("output_learn_qbc.txt", 5, 
                                        func=oracle)
\end{lstlisting}
\caption{Code example on how to start \textit{regAL} in learn mode.}
\label{fig:example_code_3}
\end{figure}

\section{Conclusion and Outlook}
Active learning shows promising possibilities to utilize machine learning methods in scenarios where computational or experimental results are scarce and expensive to obtain.
While active learning for classification problems is well understood, less attention has been given to regression problems.
To facilitate access to active learning for regression, we introduce the Python package \textit{regAL} to benchmark and perform different active learning methods for regression tasks, without specific knowledge or implementation effort from the user's side.
The package operates as a black box and requires solely a dataset to perform an experiment.
This allows users to save time on the implementation and research of various active learning methods, as well as save time by using active learning instead of traditional machine learning.\\
An experiment can be performed either as a benchmark on an already fully labeled dataset or as a learning procedure on an unlabeled dataset.
In case of a benchmark experiment, users can gain an understanding of the performance of different sample selection methods on their datasets, helping them decide whether to invest further resources in these approaches.\\
If users find active learning suitable for their scopes, they can also perform active learning on an unlabeled dataset using our package, with no need for additional implementation.
Although requiring minimal input to run an experiment, users can customize their experiment parameters, such as the training model architecture, to fit their learning scope.\\
For any experiment performed through \textit{regAL}, a detailed history of every training iteration is recorded containing all variable and fixed parameters, allowing a precise analysis of all sample selection methods.
Additionally, an experiment can be resumed after its finish or an interruption, to avoid wasting resources on retraining.\\
At the time of publication, our package performs active learning by singular sample selection.
Further developments, including potential extensions to batch-wise approaches and additional selection methods, are under consideration.\\
Finally, we have provided a demonstration of our package on the QM9 dataset and considered its runtime for different active learning methods, showing that, compared to the labeling effort, the runtime of \textit{regAL} in negligible.

\section*{Acknowledgements}
The authors acknowledge funding by the German Research Foundation (DFG, project number 512350771) and thank Nils Kramer (TU Braunschweig) for beta-testing of the \textit{regAL} package.
JP acknowledges funding from Germany’s joint federal and state program supporting early-career researchers (WISNA) established by the Federal Ministry of Education and Research (BMBF).\\

\printbibliography

\newpage
\section*{Appendix}

\end{document}